\documentclass{article}
\usepackage{graphicx}
\graphicspath{ {./images/} }

\PassOptionsToPackage{square,numbers}{natbib}

\usepackage[preprint]{neurips_2023}

\usepackage[utf8]{inputenc} 
\usepackage[T1]{fontenc}    
\usepackage{hyperref}       
\usepackage{url}            
\usepackage{booktabs}       
\usepackage{amsfonts}       
\usepackage{nicefrac}       
\usepackage{microtype}      
\usepackage{xcolor}         
\usepackage{hyperref}
\usepackage{url}
\usepackage{caption}
\usepackage{subcaption}
\usepackage{bbm}
\usepackage{graphicx}
\usepackage{resizegather}
\usepackage{amsmath,amssymb,amsfonts,algorithm,algpseudocode}

\title{Reinforcement Learning-Guided \\ Semi-Supervised Learning}

\author{%
	 Marzi Heidari \quad\; Hanping Zhang \quad\; Yuhong Guo\\[.5ex]
School of Computer Science, Carleton Univerity, Ottawa, Canada
}

\begin{document}

\maketitle

\begin{abstract}
 In recent years, semi-supervised learning (SSL) has gained significant attention due to its ability to leverage both labeled and unlabeled data to improve model performance, especially when labeled data is scarce. 
However, most current SSL methods 
rely on heuristics or predefined rules for generating pseudo-labels and leveraging unlabeled data. 
They are limited to 
exploiting loss functions and regularization methods within the standard norm.
In this paper, we propose a novel Reinforcement Learning (RL) Guided SSL method, RLGSSL, 
that formulates SSL as a one-armed bandit problem and deploys an innovative RL loss based on weighted reward
to adaptively guide the learning process of the prediction model. 
RLGSSL incorporates a carefully designed reward function 
that balances the use of labeled and unlabeled data to enhance generalization performance. 
A semi-supervised teacher-student framework is further deployed to increase the learning stability. 
We demonstrate the effectiveness of RLGSSL through extensive experiments on several benchmark datasets 
and show that our approach achieves consistent superior performance compared to state-of-the-art SSL methods. 
\end{abstract}

\section{Introduction}
Semi-supervised learning (SSL) is a significant research area in the field of machine learning, 
addressing the challenge of effectively utilizing limited labeled data alongside abundant unlabeled data. SSL techniques bridge the gap between supervised and unsupervised learning, offering a practical solution when labeling large amounts of data is prohibitively expensive or time-consuming. The primary goal of SSL is to leverage the structure and patterns present within the unlabeled data to improve the learning process, generalization capabilities, and overall performance of the prediction model. 
Over the past few years, there has been considerable interest in developing various SSL methods, and these approaches have found success in a wide range of applications, from computer vision \cite{berthelot2019mixmatch}  to natural language processing \cite{howard2018universal} 
and beyond \cite{rasmus2015semi, zhu2009introduction}.

Within the SSL domain, a range of strategies has been devised to effectively utilize the information available in both labeled and unlabeled data. Broadly, SSL approaches can be categorized into three key paradigms: regularization-based, mean-teacher-based, and pseudo-labeling methodologies. 
Regularization-based approaches form a fundamental pillar of SSL~\cite{miyato2018virtual,laine2017temporal,zhang2020consistency}. 
These methods revolve around the core idea of promoting model robustness against minor perturbations in the input data. A quintessential example in this category is Virtual Adversarial Training (VAT) \cite{miyato2018virtual}. VAT capitalizes on the introduction of adversarial perturbations to the input space, thereby ensuring the model's predictions maintain consistency. The second category, Mean-teacher based methods, encapsulates a distinct class of SSL strategies that leverage the concept of temporal ensembling. This technique aids in the stabilization of the learning process by maintaining an exponential moving average of model parameters over training iterations. Mean Teacher \cite{tarvainen2017mean} notably pioneered this paradigm with their Mean Teacher model, illustrating its efficacy across numerous benchmark tasks. Lastly, the category of Pseudo-labeling approaches has attracted attention due to its simplicity and effectiveness. These methods employ the model's own predictions on unlabeled data as ``pseudo-labels'' to augment the training process. The MixMatch \cite{berthelot2019mixmatch} framework stands as one of the leading representatives of this category, demonstrating the potential of these methods in the low-data regime.

Despite these advancements, achieving high performance with limited labeled data continues to be a significant challenge in SSL, often requiring intricate design decisions and the careful coordination of multiple loss functions. 
In this paper,
we propose to approach SSL outside the conventional design norms by 
developing a Reinforcement Learning Guided Semi-Supervised Learning (RLGSSL) method.
RL has emerged as a promising direction for addressing learning problems, with the potential to bring a fresh perspective to SSL. 
It offers a powerful framework for decision-making and optimization, which can be harnessed to discover novel and effective strategies for utilizing the information present in both labeled and unlabeled data. 
In RLGSSL, we formulate SSL as a bandit problem, 
where the prediction model serves as the policy function, and pseudo-labeling acts as the actions. 
We define a simple reward function that balances the use of labeled and unlabeled data 
and improves generalization capacity 
by leveraging linear data interpolation,
while the prediction model is trained under the standard RL framework to maximize
the empirical expected reward.
Formulating the SSL problem as such an RL task allows our approach to
dynamically adapt and respond to the data. 
Moreover, 
we further deploy a teacher-student learning framework to enhance 
the stability of learning. 
Additionally, we integrate a supervised learning loss to improve and accelerate the learning process. 
This new SSL framework has the potential to pave the way for more robust, flexible, and adaptive SSL methods.
We evaluate the proposed method through extensive experiments on benchmark datasets.
The contribution of this work can be summarized as follows:
\begin{itemize}
    \item 
We propose RLGSSL, a novel Reinforcement Learning-based approach that effectively tackles SSL 
by leveraging RL's power to learn effective strategies for generating pseudo-labels and guiding the learning process.

    \item We design a prediction assessment reward function that encourages the learning of 
	    accurate and reliable pseudo-labels 
	    while maintaining a balance between the usage of labeled and unlabeled data, thus promoting better generalization performance.
    \item
	  We introduce a novel integration framework that combines the power of 
		both RL loss and standard semi-supervised loss for SSL, 
		providing a more adaptive and data-driven approach 
that has the potential to lead to more accurate and robust SSL models.
    \item 
	    Extensive experiments demonstrate that our proposed method outperforms state-of-the-art approaches in SSL.
\end{itemize}

\section{Related Work}
\subsection{Semi-Supervised Learning}
Existing SSL approaches can be broadly classified into three primary categories: 
regularization-based methods, teacher-student-based methods, and pseudo-labeling techniques.

\textbf{Regularization-Based Methods} 
A prevalent research direction in SSL focuses on regularization-based methods, which introduce additional terms to the loss function to promote specific properties of the model. For instance, the $\Pi$-model \cite{laine2017temporal} and Temporal-Ensemble \cite{laine2017temporal} incorporate consistency regularization into the loss function, with the latter employing the exponential moving average of model predictions. Virtual Adversarial Training (VAT) \cite{miyato2018virtual} is yet another regularization-based technique that aims to make deep neural networks robust to adversarial perturbations. In a similar vein, Consistency Regularization for Generative Adversarial Networks (CR-GAN) \cite{zhang2020consistency} integrates a generative adversarial network (GAN) with a consistency regularization term, facilitating the generation of pseudo-labels for unlabeled data.

\textbf{Teacher-Student-Based Methods} 
Teacher-student-based methods offer an alternative approach in SSL research. These techniques train a student network to align its predictions with those of a teacher network on unlabeled data. 
Mean Teacher (MT) \cite{tarvainen2017mean}, a prominent example in this category, leverages an exponential moving average (EMA) on the teacher model. To enhance performance, MT + Fast SWA \cite{athiwaratkun2019there} combines Mean Teacher with Fast Stochastic Weight Averaging. Smooth Neighbors on Teacher Graphs (SNTG) \cite{luo2018smooth} takes a different approach, utilizing a graph for the teacher to regulate the distribution of features in unlabeled samples. Meanwhile, Interpolation Consistency Training (ICT) \cite{verma2022interpolation} aims to promote consistent predictions across interpolated data points by ensuring that a model's predictions on an interpolated set of unlabeled data points remain consistent with the interpolation of the predictions on those points.

\textbf{Pseudo-Labeling Methods} 
Pseudo-labeling is an effective way to
extend the labeled set when the number of labels is limited. Pseudo-Label \cite{lee2013pseudo} produces labels for unlabeled data using model predictions and filters out low-confidence predictions. MixMatch \cite{berthelot2019mixmatch} employs data augmentation to create multiple input versions, obtaining predictions for each and averaging them to generate pseudo-labels. In contrast, works such as ReMixMatch \cite{berthelot2020remixmatch}, UDA \cite{xie2020unsupervised}, and FixMatch \cite{sohn2020fixmatch} apply confidence thresholds to produce pseudo-labels for weakly augmented samples, which subsequently serve as annotations for strongly augmented samples.
Label propagation methods, including TSSDL \cite{shi2018transductive} and LPD \cite{iscen2019label}, assign pseudo-labels based on local neighborhood density. DASO \cite{oh2021distribution} combines confidence-based and density-based pseudo-labels in varying ways for each class. Approaches such as Dash \cite{xu2021dash} and FlexMatch \cite{zhang2021flexmatch} dynamically adjust confidence thresholds in a curriculum learning manner to generate pseudo-labels. Meta Pseudo-Labels \cite{pham2021meta} uses a bi-level optimization strategy, deriving the teacher update rule from student feedback, to learn from limited labeled data.
Co-Training \cite{blum1998combining} is an early representative of pseudo-lableing which involves training two classifiers on distinct subsets of unlabeled data and using confident predictions to produce pseudo-labels for one another. Similarly, Tri-Training \cite{zhou2005tri} trains three classifiers on separate unlabeled data subsets and generates pseudo-labels based on the disagreements between their predictions.

\subsection{Reinforcement Learning}

Reinforcement Learning (RL) is a field of study that focuses on optimizing an agent's decision-making abilities by maximizing the cumulative reward obtained through interactions with its environment~\citep{sutton2018reinforcement}. 
RL methodology has been widely applied to solve many other learning problems, including  
searching for optimized network architectures~\citep{zoph2016neural},
training sequence models for text generation by receiving reward signals~\citep{bahdanau2016actor,ranzato2015sequence},
and solving online planning problems~\citep{fickinger2021scalable}.
Recently, RL has been applied to fine-tune complex models that typically fail to align with users' preferences. 
Moreover, based on RL from Human Feedback (RLHF;~\citep{christiano2017deep, ziegler2019fine, stiennon2020learning}), 
ChatGPT achieves great success in dialogue generation by fine-tuning Large Language Models (LLM)
\citep{ouyang2022training}. 
It frames the training of LLM as a bandit problem, 
specifically a one-armed bandit problem~\citep{sutton2018reinforcement}, 
where the objective is to determine the optimal action (dialogue generation) for a given state (user prompt) 
within a single step. 

The bandit problem 
was originally described as a statistical decision model used by agents to optimize their decision-making process~\citep{robbins1952some}. In this problem, an agent receives a reward upon taking an action and learns to make the best decision by maximizing the given reward. The bandit problem found its application in economics and has been widely used in market learning, specifically in finding the optimal market demands or prices to maximize expected profits~\citep{rothschild1974two}. Bergemann et al.~\citep{bergemann2006bandit} and Lattimore et al.~\citep{lattimore2020bandit} have extensively discussed the literature and modern applications of the bandit problem. Additionally, Mortazavi et al.~\citep{mortazavi2022theta} introduced a Single-Step Markov Decision Process (SSMDP) to formulate the bandit problem in a manner that aligns with modern RL techniques. This advancement enables the utilization of standard RL methods on conventional bandit problems.

\begin{figure}[t]
  \centering
 \includegraphics[width= 0.78 \textwidth]{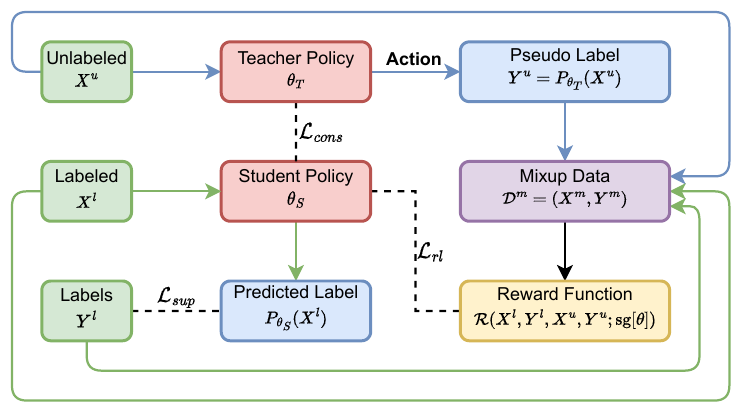} 
	 \caption{Overview of the RLGSSL Framework. 
	 The prediction networks ($\theta_S, \theta_T$) serve as the policy functions, 
	 and pseudo-labeling ($P_{\theta_T}(X^u)$) acts as the actions.
	 The model has three loss terms in total: 
	 RL loss ($\mathcal{L}_{\text{rl}}$), supervised loss ($\mathcal{L}_{\text{sup}}$), 
	 and consistency loss ($\mathcal{L}_{\text{cons}}$). 
	 The teacher policy function is used to execute the actions and compute the consistency loss,
	 while the student policy function is used for all other aspects. 
}
   \label{fig:method}
   \vskip -.1in
\end{figure}

\section{The Proposed Method}

We consider the following semi-supervised learning setting: 
the training dataset consists of a small number of labeled samples, 
$\mathcal{D}_l = (X^l, Y^l)=\{(x_i^l, {\bf y}_i^l)\}_{i=1}^{N^l}$, and a large number of unlabeled samples, 
$\mathcal{D}_u = X^u= \{x_i^u\}_{i=1}^{N^u}$, with $N^u \gg N^l$,
where $x_i^l$ (or $x_i^u$) denotes the input instance
and ${\bf y}_i^l$ denotes the one-hot label vector with length $C$. 
The goal is to train a $C$-class classifier $f_\theta: \mathcal{X} \rightarrow \mathcal{Y}$ 
that generalizes well to unseen test data drawn from the same distribution as the training data.

In this section, we present the proposed RLGSSL method, which formulates SSL 
as a one-armed bandit problem with a continuous action space,
and deploys a standard RL loss to guide the SSL process 
based on a reward function specifically designed for semi-supervised data. 
Moreover, we further incorporate a semi-supervised teacher-student framework
to augment the RL loss with a supervised loss and a prediction consistency regularization loss, 
aiming to enhance the learning stability and efficacy. 
Figure \ref{fig:method} illustrates 
the overall framework of the proposed RLGSSL,
and the following subsections will elaborate on the approach. 

\subsection{Reinforcement Learning Formulation for SSL}
\label{sec:RLGSSL}

We treat SSL as a special one-armed bandit problem with a continuous action space. 
One-armed bandit problem can be considered a single-step Markov Decision Process (MDP)~\citep{mortazavi2022theta}. 
In this problem, the agent takes a single action and receives a reward based on that action. 
The state of the environment is not affected by the action. 
The one-armed bandit problem involves selecting an action to maximize an immediate reward,
which can be regarded as learning a policy function under the RL framework. 
Formulating SSL as a one-armed bandit problem within the RL framework
and deploying RL techniques to guide SSL
requires defining the following key components: state space $\mathcal{S}$, 
action space $\mathcal{A}$, 
a policy function $\pi: \mathcal{S} \to \mathcal{A}$, and  
a reward function $\mathcal{R}: \mathcal{S}\times \mathcal{A}\to \mathbb{R}$.
The objective is to learn an optimal policy $\pi^\star$ 
that maximizes the one-time reward $\mathcal{R}(s, \pi(\cdot|s))$ 
in the given environment (state $s$): 
$\pi^\star=\arg\max_\pi\; J_r(\pi)=\mathbb{E}_{\pi}[\mathcal{R}(s,\pi(\cdot|s))]$.

\paragraph{State}
The state encapsulates the provided knowledge about the environment and is used as input for the policy function.
As the action does not affect the state of the environment under one-armed bandit problem, 
we use the observed data from the SSL problem as the state; i.e., $s=(X^l, Y^l, X^u)$. 

\paragraph{Action and Policy Function}
As the goal of SSL is to learn an optimal classifier $f_\theta$ (i.e., prediction network parameterized with $\theta$), 
we use the classifier $f_\theta$, usually denoted by its parameters $\theta$, as the policy function $\pi_\theta$
to {\em unify the goals of RL and SSL}. 
In particular, we consider a probabilistic policy function/prediction network 
$\pi_\theta(\cdot)=P_\theta(\cdot)$. 
Since policy function is used to project a mapping from the state $s$ to the action space, 
$a=\pi_\theta(\cdot|s)$, 
by using a probabilistic prediction network as the policy function, 
it naturally determines a continuous action space $\mathcal{A}$.
Specifically, given the fixed state $s$, taking an action
is equivalent to making probabilistic predictions on the unlabeled data in $s$:
$Y^u=P_\theta(X^u)=\pi_\theta(\cdot|s)$, as the labeled data already has labels.  
For each unlabeled instance $x_i^u$, 
the action is a probability vector produced as ${\bf y}_i^u=P_\theta(x_i^u)$,
which can be regarded as soft pseudo-labels in the SSL setting.
This links the action of RL to the pseudo-labeling in SSL. 
%
\subsubsection{Reward Function}
The reward function serves as feedback to evaluate the performance of the action (prediction) provided by the policy. It needs to be thoughtfully crafted to maximize the model's ability to extract useful information from both labeled and unlabeled data, which is central to the SSL paradigm. 
The underlying motivation is to guide the learning process to induce a more generalizable and robust prediction model. 
To this end, we adopt a data mixup~\cite{zhang2018mixup}
 strategy to produce new data points 
from the given labeled data $(X^l,Y^l)$ and pseudo-labeled data $(X^u,Y^u)$,
which together form $(s,a)$,
through linear data interpolation,  
and assess the prediction model's generalization ability on such data points as the reward signal. 
This decision is inspired by the proven effectiveness of mixup in enhancing model performance in various tasks. 
The idea of data mix-up is to generate virtual training examples by creating convex combinations of pairs of input data and their corresponding labels. This technique encourages the model to learn more fluid decision boundaries, leading to improved generalization capabilities.

Specifically, we propose to generate new data points by performing {\em inter-mixup} 
between labeled and unlabeled data points,
aiming to maintain a balanced utilization of both labeled and unlabeled data. 
In order to address the size discrepancy between the labeled dataset $\mathcal{D}^l$ and 
the unlabeled dataset $\mathcal{D}^u$, with $N^u \gg N^l$, 
we replicate the labeled dataset $\mathcal{D}^l$ by a factor of 
$r = \lceil\frac{N^u}{N^l}\rceil$ times, 
resulting in an extended labeled dataset $\widetilde{\mathcal{D}}^l$.
After shuffling the data points in each set, 
we generate a mixup data point by mixing an unlabeled point 
$x_i^u \in \mathcal{D}^u$ with a labeled point $x_i^l \in \widetilde{\mathcal{D}}^l$ 
along with their corresponding pseudo-label ${\mathbf{y}^u_i} \in \mathcal{D}^u$ and 
label ${\mathbf{y}^l_i} \in \widetilde{\mathcal{D}}^l$:
\begin{equation}
\begin{gathered}
x_i^{\text{m}}=\mu\,x_i^u+(1-\mu)\,x^l_i,\qquad 
{\mathbf{y}}_i^{\text{m}}=\mu\,{\mathbf{y}}_i^u + (1-\mu)\,{\mathbf{y}}_i^l
\end{gathered}
\label{eq:mix}
\end{equation}
where the mixing parameter $\mu$ is sampled from a Beta distribution.
With this procedure, we can generate 
$N^{\text{m}}=N^u$ mixup samples
by mixing all the unlabeled data with the extended labeled data.  

We then define the reward function to measure the negative mean squared error (MSE) 
between the model's prediction $P_\theta(x_i^{\text{m}})$ and the mixup label $\mathbf{y}_i^{\text{m}}$ 
for each instance in the mixup set. 
This results in a single, comprehensive metric that quantifies the overall (negative) 
disagreement between the model's predictions and the mixup labels over a large set of interpolated data points:
\begin{equation}
\mathcal{R}(s, a;\text{sg}[\theta]) =
\mathcal{R}(X^l, Y^l, X^u, Y^u;\text{sg}[\theta]) 
	= - \frac{1 }{C\cdot N^\text{m} }\sum\nolimits_{i=1}^{N^\text{m}} ||P_{\theta}(x_i^{\text{m}})-\mathbf{y}_i^{\text{m}}||^2_2
\label{eq:reward}
\end{equation}
where $C$ denotes the number of classes and $\text{sg}[\cdot]$ is the stop gradient operator 
which stops the flow of gradients during the backpropagation process. 
This ensures that the reward function is solely employed for model assessment, 
rather than being directly utilized for model updating, 
enforcing the working mechanisms of RL. 
Mixup labels capture both the supervision information in the labeled data
and the uncertainty in the pseudo-labels of unlabeled data. 
With the designed reward function, a good reward value can only be returned 
when the prediction model not only exhibits strong alignment with the labeled data 
but also delivers accurate predictions on the unlabeled data. 
Consequently, through RL, this reward function will 
not only promote accurate predictions but also enhance the model's robustness and generalizability.

\subsubsection{Reinforcement Learning Loss}

By deploying the probabilistic pedictions on the unlabeled data, 
$Y^u= \pi_\theta(X^u)=P_\theta(X^u)$, 
as the action, we indeed adopt a deterministic policy. 
Following the principle of one-armed bandit problem on 
maximizing the expected one-time reward w.r.t. the policy output, 
we introduce a weighted negative reward based on the deterministic policy output
as the RL loss for the proposed RLGSSL.
Specifically, we treat the output of the policy network, $Y^u$, 
as a uniform distribution over the set of $N^u$ probability vectors, 
$\{{\bf y}_1^u,\cdots, {\bf y}_{N^u}^u\}$,
predicted for the unlabeled instances.
Let ${\bf e}={\bf 1}/C$ denote a discrete uniform distribution vector with length $C$. 
We design the following KL-divergence weighted negative reward as the RL loss:
\begin{equation}
\begin{aligned}
\label{eqa:rl_loss-kl}
\mathcal{L}_{\text{rl}} 
&= - \mathbb{E}_{{\bf y}_i^u\sim\pi_\theta} \mbox{KL}({\bf e}, {\bf y}_i^u)\mathcal{R}(s,a;\text{sg}[\theta])\\ 
&= - \mathbb{E}_{x_i^u\in \mathcal{D}_u} \mbox{KL}({\bf e}, P_\theta(x_i^u))\mathcal{R}(s,a;\text{sg}[\theta])\\ 
\end{aligned}
\end{equation}
where the KL-divergence term measures the distance of each label prediction probability vector
${\bf y}_i^u$ from a uniform distribution vector.
Given that a uniform probability distribution signifies the least informative prediction outcome, 
the expected KL-divergence captures the level of informativeness in the policy output
and hence serves as a meaningful weight for the reward, 
which inherently encourages the predictions to exhibit greater discrimination.

The minimization of this loss function over the prediction network parameterized by $\theta$ 
is equivalent to learning an optimal policy function $\pi_\theta$ by maximizing 
the KL-divergence weighted reward,
which aims at an optimal policy function (also the probabilistic classifier $P_\theta$) 
that not only maximizes the reward signal but is also discriminative.
From the perspective of SSL, the utilization of this novel RL loss introduces a fresh approach 
to designing prediction loss functions. 
Instead of directly optimizing the alignment between predictions and targets, 
it offers a gradual learning process guided by reward signals. 
This innovative approach presents a more 
adaptive and flexible 
solution for complex data scenarios,
where the traditional optimization-based methods may fall short.
%

\subsection{Teacher-Student Framework for RL-Guided SSL}
\label{sec:teacher-student}

Teacher-student models~\cite{tarvainen2017mean} 
have been popularly deployed to exploit unlabeled data for SSL,
improving the learning stability.
We extend this mechanism to provide a teacher-student framework
for RL-guided SSL 
by maintaining a dual set of model parameters: 
the student policy/model parameters $\theta_S$, and the teacher policy/model parameters $\theta_T$. 
The student model is directly updated through training, 
whereas the teacher model is updated via an exponential moving average (EMA) of the student model. 
The update is conducted as follows:
\begin{equation}
\theta_T= \beta\,\theta_T + (1-\beta)\,\theta_S
\label{eq:ema}
\end{equation}
where $\beta$ denotes a hyperparameter that modulates the EMA's decay rate. 
The utilization of the EMA update method ensures a stable and smooth transfer of knowledge from the student model 
to the teacher model. 
This leads to a teacher model with consistent and reliable parameter values 
that are not susceptible to random or erratic fluctuations during the training process.
Leveraging this desirable characteristic, 
we propose to {\em employ the teacher model for executing actions} within the RL framework 
described in the subsection \ref{sec:RLGSSL} above; 
that is, $Y^u=P_{\theta_T}(X^u)$, 
while retaining the student model for other aspects. 
By doing so, we ensure that stable actions are taken, 
reducing the impact of random noise in the generated pseudo-labels and enhancing the accuracy of reward evaluation.

Within the teacher-student framework, we further propose to augment the RL loss 
with a supervised loss $\mathcal{L}^{\text{sup}}$ on the labeled data 
and a consistency regularization loss $\mathcal{L}^{\text{cons}}$ on the unlabeled data. 
We adopt a standard cross-entropy loss function $\ell_{CE}$ to compute the supervised loss,
promoting accurate predictions on $\mathcal{D}^l$
where the ground-truth labels are available: 
\begin{equation}
\label{eq:cl-loss}
\mathcal{L}^{\text{sup}} = 
	\mathbb{E}_{(x^{l}, {\bf y}^{l})\in \mathcal{D}^l}
	\left[\ell_{CE}\left(P_{\theta_{S}}(x^{l}),{\bf y}^{l}\right) \right]
\end{equation}
This loss can enhance effective exploitation of the ground-truth label information, 
providing a solid basis for exploring the parameter space via RL. 
The consistency loss $\mathcal{L}^{\text{cons}}$ is deployed to encourage 
the prediction consistency between the student and teacher models on the unlabeled data $\mathcal{D}^u$:
\begin{equation}
\label{eq:cons-loss}
\mathcal{L}^{\text{cons}} = 
	\mathbb{E}_{x^u\in\mathcal{D}^u} \left[\ell_{\text{KL}}\left(P_{\theta_{S}}(x^u), P_{\theta_T}(x^u)\right)\right]
\end{equation}
where $\ell_{\text{KL}}(\cdot,\cdot)$ denotes the Kullback-Leibler divergence 
between two probability distributions.
By enforcing consistency, this loss encourages the student model to make more confident and reliable predictions, 
reducing the impact of random or misleading information in the training set.
It also acts as a form of regularization, discouraging the student model from overfitting to the labeled data.

\begin{algorithm}[t]
  \caption{Pseudo-Label Based Policy Gradient Descent}
  \begin{algorithmic}
	  \State{{\bf Input:} $\mathcal{D}^l, \mathcal{D}^u$, and extended  $\widetilde{\mathcal{D}}^l$;
	  \quad initialized $\theta_S, \theta_T$;\quad hyperparameters}
	  \vskip .01in
\For{$iteration=1$ to maxiters}
\For{${x}_i^u \in \mathcal{D}^u$}
	  \State Compute soft pseudo-label vector ${\bf y}_i^u = P_{\theta_T}({x}_i^u)$ to form 
	  $({x}_i^u, {\bf y}_i^u)$
\EndFor
\State Generate mixup data $\mathcal{D}^m$ =$(X^{m},Y^{m})$ on $\mathcal{D}^u$ and $\widetilde{\mathcal{D}}^l$ 
	  using Eq.(\ref{eq:mix}) with shuffling
\For{$step=1$ to maxsteps} 
	  \State Draw a batch of data $B=\{({x}_i^m, {\bf y}_i^m)\}$ from $\mathcal{D}^m$
	  \State Calculate the reward function $\mathcal{R}(\cdot;sg[\theta_S])$ using the batch $B$
	  \State Compute the objective in Eq.(\ref{eq:total-loss})
\State Update the policy parameters $\theta_S$ via gradient descent 
	  \State Update teacher model $\theta_T$ via EMA in Eq.(\ref{eq:ema})
\EndFor
\EndFor
\end{algorithmic}
\label{alg:batch-wise2}
\end{algorithm}

\subsection{Training Algorithm for RL-Guided SSL}
\label{sec:optimization-process}

The learning objective for the RLGSSL approach is formed by combining the reinforcement learning loss
$\mathcal{L}_\text{rl}$ with the two augmenting loss terms, 
the supervised loss $\mathcal{L}_\text{sup}$ and the consistency loss $\mathcal{L}_{\text{cons}}$,
using hyperparameters $\lambda_1$ and $\lambda_2$:
\begin{equation}
\label{eq:total-loss}
\mathcal{L}(\theta_{S}) =\mathcal{L}_{\text{rl}} + \lambda_1 \mathcal{L}_{\text{sup}} +\lambda_2 \mathcal{L}_{\text{cons}}
\end{equation}
By deploying such a joint loss,
the RLGSSL framework can benefit from the strengths of 
both reinforcement exploration and semi-supervised learning. 
The RL component, in particular, introduces a dynamic aspect to the learning process, enabling the model to improve iteratively based on its own experiences. This innovative combination of losses allows the model to effectively learn from limited labeled data while still exploiting the abundance of unlabeled data.

We develop a stochastic batch-wise gradient descent algorithm to minimize the joint objective in Eq.(\ref{eq:total-loss})
for RL-guided semi-supervised training. 
The algorithm is summarized in Algorithm~\ref{alg:batch-wise2}.

\section{Experiments}
\subsection{Experimetal Setup}

\paragraph{Datasets}
We conducted comprehensive experiments on three 
image classification benchmarks: CIFAR-10, CIFAR-100 \cite{krizhevsky2009learning}, and SVHN \cite{netzer2011reading}. 
We adhere to the conventional dataset splits used in the literature.
CIFAR-10 and CIFAR-100 are split into 50,000 training images and 10,000 test images.
The SVHN dataset
provides 73,257 images for training and 26,032 images for testing.
Consistent with previous works, we preserved the labels of a randomly selected subset of training samples with an equal number of samples for each class in CIFAR and SVHN, and left the remaining samples unlabeled. 
We performed experiments on CIFAR-10 with 4000, 2000, and 1000 labeled samples, on CIFAR-100 with 10000 and 4000 labeled samples, and on SVHN with 1000 and 500 labeled samples.

\paragraph{Implementation Details}  
In our study, we adopted the data augmentation strategy from previous works \cite{luo2018smooth,tarvainen2017mean} by applying random $2 \times 2$ translations to the training set images.  
We conducted experiments using two different
network architectures: a 13-layer Convolutional Neural Network (CNN-13) and a Wide-Residual Network with 28 layers and a widening factor of 2 (WRN-28). For training the CNN-13 architecture, we employed the Stochastic Gradient Descent (SGD) optimizer with a Nesterov momentum of 0.9. We used an L2 regularization coefficient of 1e-4 for the CIFAR-10 and CIFAR-100 datasets, and 5e-5 for the SVHN dataset. The initial learning rate 
was set to 0.1. To schedule the learning rate effectively, we utilized the cosine learning rate annealing technique as proposed in \cite{loshchilov2017sgdr,verma2022interpolation}. For the WRN-28 architecture, we followed the suggestion from \cite{berthelot2019mixmatch} and used an L2 regularization coefficient of 4e-4. 
To compute the parameters of the teacher model, 
we employed the EMA method with a decay rate $\beta$ of 0.999. 
We selected all hyperparameters and training techniques based on relevant studies to ensure a fair comparison between our approach and the existing methods. 
Specifically for RLGSSL, we set the batch size to 128, $\lambda_1$ to 0.1, and $\lambda_2$ to 0.1. Before beginning with RLGSSL, we pre-train the model for 50 epochs using the Mean-Teacher algorithm. We then proceed to train RLGSSL for 400 epochs. We conducted five runs of the experiments and reported the mean test errors with their standard deviations.

\begin{table}[t]
\centering
\caption{ Performance of RLGSSL and state-of-the-art SSL algorithms with the CNN-13 network. We report
the average test errors and the standard deviations of 5 trials.}
\vskip .05in
\label{table:cifar}
\resizebox{\textwidth}{!}{
\begin{tabular}{l|c|c|c|c|c}
\hline
Dataset                   & \multicolumn{3}{c|}{CIFAR-10}                        & \multicolumn{2}{c}{CIFAR-100}                                                              \\ 
Number of Labeled Samples & {1000}           & {2000}           & {4000}           & {4000}           & 10000          \\
\hline
Supervised & ${39.95}_{(0.75)}$ & ${27.67}_{(0.12)}$ & ${20.42}_{(0.21)}$ & ${58.31}_{(0.89)}$ & ${44.56}_{(0.30)}$ \\
Supervised + MixUp \cite{zhang2018mixup} & ${31.83}_{(0.65)}$ & ${24.22}_{(0.15)}$ & ${17.37}_{(0.35)}$ & ${54.87}_{(0.07)}$ & ${40.97}_{(0.47)}$ \\
$\Pi$-model \cite{laine2017temporal} & ${28.74}_{(0.48)}$ & ${17.57}_{(0.44)}$ & ${12.36}_{(0.17)}$ & ${55.39}_{(0.55)}$ & ${38.06}_{(0.37)}$ \\
Temp-ensemble \cite{laine2017temporal} & ${25.15}_{(1.46)}$ & ${15.78}_{(0.44)}$ & ${11.90}_{(0.25)}$ & {-} & ${38.65}_{(0.51)}$ \\
Mean Teacher\cite{tarvainen2017mean} & ${21.55}_{(0.53)}$ & ${15.73}_{(0.31)}$ & ${12.31}_{(0.28)}$ & ${45.36}_{(0.49)}$ & ${35.96}_{(0.77)}$ \\
VAT \cite{miyato2018virtual} & ${18.12}_{(0.82)}$ & ${13.93}_{(0.33)}$ & ${11.10}_{(0.24)}$ & {-} &{-} \\
SNTG \cite{luo2018smooth} & ${18.41}_{(0.52)}$ & ${13.64}_{(0.32)}$ & ${10.93}_{(0.14)}$ & {-} & ${37.97}_{(0.29)}$ \\
Learning to Reweight \cite{ren2018learning} & ${11.74}_{(0.12)}$ & -& ${9.44}_{(0.17)}$ & ${46.62}_{(0.29)}$ & ${37.31}_{(0.47)}$ \\
MT + Fast SWA \cite{athiwaratkun2019there} & ${15.58}$ & ${11.02}$ & ${9.05}$ & {-} & ${33.62}_{(0.54)}$ \\
ICT \cite{verma2022interpolation} & ${12.44}_{(0.57)}$ & ${8.69}_{(0.15)}$ & ${7.18}_{(0.24)}$ & ${40.07}_{(0.38)}$ & ${32.24}_{(0.16)}$ \\
RLGSSL (Ours) & $\mathbf{9.15}_{(0.57)}$ & $\mathbf{6.90}_{(0.11)}$ & $\mathbf{6.11}_{(0.10)}$ &$\mathbf{36.92}_{(0.45)}$& $\mathbf{29.12}_{(0.20)}$ \\
\hline
\end{tabular}}
\vskip -.05in
\end{table}

\begin{table}[t]
\vskip -.05in
\centering
\caption{Performance of RLGSSL and state-of-the-art SSL algorithms with the CNN-13 network. We report the average test errors and the standard deviations of 5 trials.}
\vskip .05in
\label{table:svhn}
\setlength{\tabcolsep}{2pt}
\resizebox{\textwidth}{!}{
\begin{tabular}{l|c|c|c|c|c|c|c}
\hline
 & VAT \cite{miyato2018virtual} & $\Pi$-model \cite{laine2017temporal} & Temp-ensemble \cite{laine2017temporal} & MT \cite{tarvainen2017mean} & ICT \cite{verma2022interpolation} & SNTG \cite{luo2018smooth} & RLGSSL (Ours)\\
\hline
SVHN/500 & {-} & ${6.65}_{(0.53)}$ & ${5.12}_{(0.13)}$ & ${4.18}_{(0.27)}$ & ${4.23}_{(0.15)}$ & ${3.99}_{(0.24)}$ & $\mathbf{3.12}_{(0.07)}$ \\
SVHN/1000 & ${5.42}_{(0.00)}$ & ${4.82}_{(0.17)}$ & ${4.42}_{(0.16)}$ & ${3.95}_{(0.19)}$ & ${3.89}_{(0.04)}$ & ${3.86}_{(0.27)}$ & $\mathbf{3.05}_{(0.04)}$ \\
\hline
\end{tabular}}
\vskip -.05in
\end{table}


\begin{table}[t]
\caption{Performance of RLGSSL with the WRN-28 network. Average test errors and standard deviations of
5 trials are reported.}
\vskip .05in
\label{table:wrn}
\resizebox{\textwidth}{!}{
\begin{tabular}{l|c|c|c|c|c}
\hline
{Dataset} & \multicolumn{3}{|c|}{CIFAR-10}                        & \multicolumn{1}{c}{CIFAR-100}&\multicolumn{1}{|c}{SVHN} \\
Number of Labeled Samples                                         & 1000     & 2000     & 4000     & 10000  &1000  \\
\hline
Mean Teacher \cite{tarvainen2017mean}& $17.32_{(4.00)}$ & $12.17_{(0.22)}$ & $10.36_{(0.25)}$ & - &$5.65_{(0.45)}$ \\
VAT \cite{miyato2018virtual} & $18.68_{(0.40)}$ & $14.40_{(0.15)}$ & $11.05_{(0.31)}$ & - &${5.35_{(0.19)}}$\\
MixMatch\cite{berthelot2019mixmatch} & $7.75_{(0.32)}$ & $7.03_{(0.15)}$ & $6.24_{(0.06)}$ & $30.84_{(0.29)}$ &$3.27_{(0.31)}$\\
Meta Pseudo-Labels \cite{pham2021meta} & - & - & $3.89_{(0.07)}$ & -&$1.99_{(0.07)}$\\
RLGSSL (Ours) & $\mathbf{4.92}_{(0.25)}$ & $\mathbf{4.24}_{(0.10)}$ & $\mathbf{3.52}_{(0.06)}$ & $\mathbf{28.82}_{(0.22)}$&$\mathbf{1.92}_{(0.05)}$\\
\hline
\end{tabular}}
\end{table}

\begin{table}[t]
\vskip -.05in
\centering
\caption{Ablation study results. We report the test errors on CIFAR-100 with 10000, and 4000 labels on  CNN-13 backbone 5 trials.}
\vskip .05in
\label{table:main-ablation}
\setlength{\tabcolsep}{4pt}
\resizebox{\textwidth}{!}{\begin{tabular}{l|c|c|c|c|c|c}
\hline
 & RLGSSL & $- \text{w/o } \mathcal{L}_\text{rl}$ & $- \text{w/o } \mathcal{L}_\text{sup}$ & $- \text{w/o } \mathcal{L}_\text{cons}$ & $- \text{w/o }$ EMA & $- \text{w/o }$ mixup\\
\hline
CIFAR-100/4000 & $\mathbf{36.92}_{(0.45)}$ & ${44.92}_{(0.55)}$ & ${39.52}_{(0.58)}$ & ${38.78}_{(0.48)}$ & ${43.12}_{(0.52)}$ & ${40.12}_{(0.51)}$ \\
CIFAR-100/10000 & $\mathbf{29.12}_{(0.20)}$ & ${33.12}_{(0.52)}$ & ${32.67}_{(0.45)}$ & ${31.48}_{(0.32)}$ & ${32.84}_{(0.45)}$ & ${31.48}_{(0.32)}$   \\
\hline
	& RLGSSL & $\mathcal{R}=1 $ & $\mathcal{R}:\mu=0$ & $\mathcal{R}(L_2\rightarrow\mbox{KL})$ & $\mathcal{R}(L_2\rightarrow\mbox{JS})$  & $\mathcal{R}:$ w/o sg[$\theta$] \\
\hline
CIFAR-100/4000 & $\mathbf{36.92}_{(0.45)}$ & ${39.52}_{(0.63)}$ & ${39.54}_{(0.33)}$ & ${38.02}_{(0.42)}$ & ${39.52}_{(0.45)}$ & ${40.62}_{(0.55)}$ \\
CIFAR-100/10000 & $\mathbf{29.12}_{(0.20)}$ & ${31.25}_{(0.62)}$ & ${32.37}_{(0.57)}$ & ${31.12}_{(0.52)}$ & ${31.39}_{(0.68)}$ & ${32.12}_{(0.62)}$   \\
\hline
\end{tabular}}
\vskip -.1in
\end{table}

\subsection{Comparison Results}
We compare RLGSSL with 
a great number of SSL algorithms, 
including 
Supervised + MixUp \cite{zhang2018mixup}, $\Pi$-model \cite{laine2017temporal}, Temp-ensemble \cite{laine2017temporal}, Mean Teacher \cite{tarvainen2017mean}, VAT \cite{miyato2018virtual}, SNTG \cite{luo2018smooth}, Learning to Reweight \cite{ren2018learning}, MT + Fast SWA \cite{athiwaratkun2019there}, MixMatch \cite{berthelot2019mixmatch}, Meta Pseudo-Labels \cite{pham2021meta},  ICT \cite{verma2022interpolation}, using CNN-13 or WRN-28 as the backbone network. 

Table \ref{table:cifar} reports the comparison results on CIFAR-10 with 4000, 2000, and 1000 labeled samples and on CIFAR-100 with 10000 and 4000 labeled samples when CNN-13 is used as the backbone network.
For CIFAR-10, RLGSSL outperforms the compared methods across all 
settings with different numbers of labeled samples, i.e., 1000, 2000, and 4000. 
In terms of CIFAR-10 performance with 1000 labeled samples, RLGSSL surpasses ICT, the second best method, by a significant margin of $3.29\%$, achieving an average test error of $9.15$. This pattern of outperformance continues with 2000 and 4000 labeled samples, where RLGSSL yields lower test error rates compared to ICT with a margin of $1.79\%$ and $1.07\%$  respectively. The results on the CIFAR-100 dataset are similarly impressive. For 4000 labeled samples, RLGSSL again outperforms ICT,
the second best method, with a margin  of $3.15\%$. As the number of labeled samples escalates to 10,000, RLGSSL maintains its performance edge, achieving a test error that outperforms ICT by a notable margin of $3.12\%$.

Table \ref{table:svhn} reports the comparison results on the SVHN dataset with CNN-13 as the backbone network. 
Our method, RLGSSL, surpasses all other SSL techniques for both settings. 
Specifically, for 500 labeled samples, RLGSSL achieves a test error of $3.12\%$, which is $0.87\%$ lower than the second-best method,  SNTG. For 1000 labeled samples, RLGSSL continues to show superior performance with a test error of $3.05\%$, outperforming the second-best method,  SNTG, by $0.81\%$.

Table \ref{table:wrn} reports the performance of the proposed RLGSSL method with the WRN-28 network on different datasets.
On the CIFAR-10 dataset, with the number of labeled samples increasing from 1000 to 4000, RLGSSL showed consistently better performance compared to other methods. For 1000 labeled samples, our method improved over the best competing method, MixMatch, by a margin of $2.83\%$. This margin is $2.79\%$  for 2000 samples and  $2.72\%$ for 4000 samples. In the CIFAR-100 dataset, with 10,000 labeled samples, RLGSSL surpassed the MixMatch method by a significant margin of $2.02\%$. On the SVHN dataset with 1000 labeled samples, our RLGSSL outperformed the Meta Pseudo-Labels. This confirms the robustness of RLGSSL across various settings, even when the amount of labeled data is limited.

These results show the effectiveness of RLGSSL, demonstrating its consistent superior performance across different numbers of labeled samples. Furthermore, RLGSSL's dominance across both CIFAR and SVHN datasets substantiates its robustness and adaptability to different datasets and scenarios.

\subsection{Ablation Study}

In order to evaluate the significance of various components of our RLGSSL approach, we executed an ablation study on the CIFAR-100 dataset using the CNN-13 network. 
In particular, we compared the full model RLGSSL with the following variants: 
(1) “$\; - \text{w/o } \mathcal{L}_\text{rl}$”, which does not incorporate the Reinforcement Learning (RL) loss term; 
(2) “ $\; - \text{w/o } \mathcal{L}_\text{sup}$”, which excludes 
the supervised loss term; 
(3) “$\; - \text{w/o } \mathcal{L}_\text{cons}$”, which  does not include the consistency loss term; 
(4) “$\; - \text{w/o }$ EMA”, which drops the teacher model 
by disabling the EMA update; 
and 
(5) “$-\text{w/o }$ mixup”, which only uses unlabelled data in the reward function ($\mu=1$), and the mixup operation is excluded.
The ablation results are reported in the top section of Table \ref{table:main-ablation}.
The full model, RLGSSL, achieved the lowest test error, confirming the overall effectiveness of our method. The most significant observation from our study lies in the removal of the RL loss, denoted as $\mathcal{L}_{\text{rl}}$. Upon removal of this component, we witness a substantial increase in test errors, which highlights the indispensable role played by the RL component in our model.  The ablation study further illustrates the importance of each component 
by analyzing the performance of the model when the component is removed. 
In each of these cases, we observe that the removal of any individual component consistently leads to an increase in test errors. This finding underpins the notion that each component of the RLGSSL model plays a significant role in the overall performance of the model.

In addition, we also conducted another set of ablation study centered on the proposed RL loss and the reward function.
We compared the full model RLGSSL with the following variants: 
(1) "$\mathcal{R}=1$", which drops RL by setting the reward as a constant 1;   
(2) $\mathcal{R}:\mu=0$, which only uses labeled data to compute the reward by setting $\mu=0$;   
(3) $\mathcal{R}(L_2\rightarrow\mbox{KL})$, 
which replaces the squared L2 norm distance in the reward function in Eq.(\ref{eq:reward}) 
with the KL-divergence distance; 
(4) $\mathcal{R}(L_2\rightarrow\mbox{JS})$,    
which replaces the squared L2 norm distance in the reward function in Eq.(\ref{eq:reward}) 
with the JS-divergence distance; 
and 
(5) $\mathcal{R}:$ w/o sg[$\theta$], which removes the stop-gradient operator from the reward function
and makes the reward function differentiable w.r.t $\theta$.
The ablation results are reported in the bottom section of Table \ref{table:main-ablation}.
We can see that all these variants with altered reward functions 
produced degraded performance comparing to the full model
with the proposed reward function. 
In particular, the performance degradation of "$\mathcal{R}=1$" and "$\mathcal{R}:$ w/o sg[$\theta$]" 
that drop RL in different manners
further validates the contribution of the proposed framework of guiding SSL with RL.

\subsection{Hyperparameter Analysis}
\begin{figure}[t]
\centering
\begin{subfigure}{0.40\textwidth}
\centering
\includegraphics[width = .9\textwidth, height=1.25in]{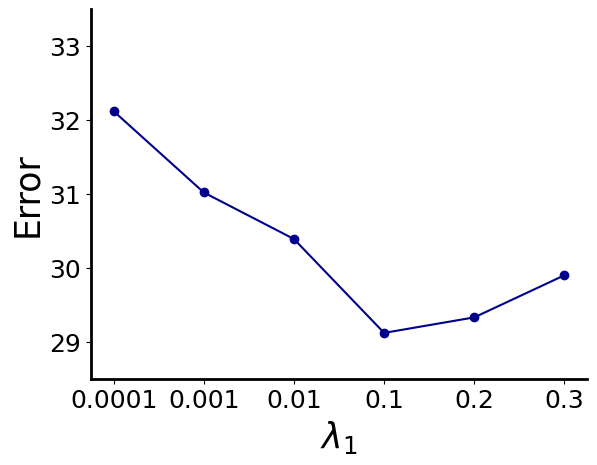}
\caption{$\lambda_1$}
\end{subfigure}\qquad\quad
\begin{subfigure}{0.40\textwidth}
\centering
\includegraphics[width = .9\textwidth, height=1.25in]{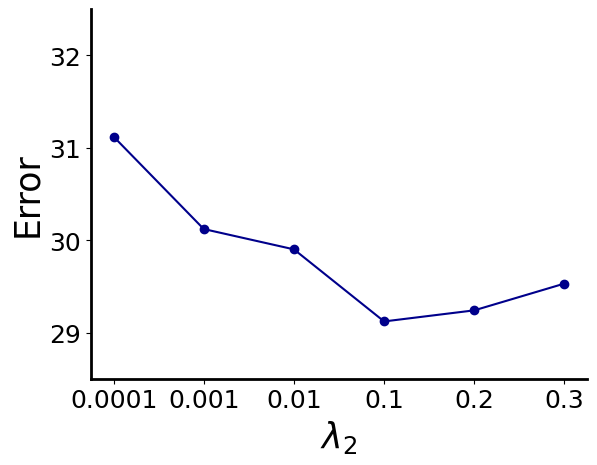}
\caption{$\lambda_2$}
\end{subfigure}
\caption{Sensitivity analysis for four hyperparameters $\lambda_1$ and  $\lambda_2$ CIFAR-100 using 10000 labeled samples (a) $\lambda_1$, (b) $\lambda_2$.}
\label{fig:hyper_sen}
 \vskip -.15in
\end{figure}
We conduct sensitivity analysis over the two hyperparameters of the proposed RLGSSL: 
$\lambda_1$---the trade-off parameter for the supervised loss, 
and $\lambda_2$---the trade-off parameter for the consistency loss. 

The results are reported in Figure \ref{fig:hyper_sen}.
In the case of $\lambda_1$, lower values (e.g., $1e-4$ and $1e-3$) result in less emphasis on the supervised loss term in the objective function. As a result, the model might not learn effectively from the limited available labeled data, 
leading to increased test error rates. 
Conversely, higher values of $\lambda_1$ (e.g., $0.2$ and $0.3$) may overemphasize the supervised loss term, 
potentially causing the model to overfit to the labeled data and ignore useful information from the unlabeled data. 
The sweet spot lies in the middle (around $0.1$), where the model strikes a balance between learning from labeled data and leveraging information from unlabeled data.

Regarding $\lambda_2$, very low values (e.g., $1e-4$ and $1e-3$) may not enforce sufficient consistency in the model predictions on unlabeled data, resulting in a model that fails to generalize well. 
However, if $\lambda_2$ is too high (e.g., $0.2$ and $0.3$), the model may overemphasize the consistency constraint, possibly leading to a model that is too rigid to capture the diverse patterns in the data. An optimal value of $\lambda_2$ (around $0.1$ in our experiments) ensures a good balance between encouraging prediction consistency and allowing the model to adapt to the diverse patterns in the data.

The optimal value choice for hyperparameters $\lambda_1$ and $\lambda_2$ (around 0.1) 
also validates that the RL loss is the main leading term, while the supervised loss and consistency loss are augmenting terms.


\section{Conclusion}
In this paper, we presented Reinforcement Learning-Guided Semi-Supervised Learning (RLGSSL), a unique approach that integrates the principles of RL to tackle the challenges inherent in SSL. 
This initiative was largely driven by the limitations of conventional SSL techniques. 
RLGSSL employs a distinctive strategy where an RL-optimized reward function is utilized. 
This function adaptively promotes better generalization performance through
more effectively leveraging both labeled and unlabeled data.
We also further incorporated a student-teacher framework to integrate the strengths of RL and SSL. 
Extensive evaluations were conducted on multiple benchmark datasets,
comparing RLGSSL to existing state-of-the-art SSL techniques. 
RLGSSL consistently outperformed these other techniques across all the datasets, which attests to the effectiveness and generalizability of our approach. The results underline the potential of integrating RL principles into SSL, and the RLGSSL method introduced in this paper is a significant stride in this direction.

\bibliographystyle{ieeetr}
\bibliography{neurips_2023}

\end{document}